\documentclass[journal]{IEEEtran}

\usepackage{cite}
\usepackage{amsmath,amssymb,amsfonts}
\usepackage{algorithm}
\usepackage{algpseudocode}
\usepackage{graphicx}
\usepackage{caption}
\usepackage{subcaption}
\usepackage[compact]{titlesec}
\usepackage{textcomp}
\usepackage{xcolor}
\usepackage{hyperref}
\usepackage{titlesec}
\usepackage{subcaption}

\titlespacing{\section}{2pt}{2pt}{2pt}

\title{From Federated Learning to Quantum Federated Learning for Space-Air-Ground Integrated Networks}
\author{
    \IEEEauthorblockN{
        Vu Khanh Quy\IEEEauthorrefmark{1},
        Nguyen Minh Quy\IEEEauthorrefmark{1}, 
        Tran Thi Hoai\IEEEauthorrefmark{2},
        Shaba Shaon\IEEEauthorrefmark{3}, 
        Md Raihan Uddin\IEEEauthorrefmark{3},
        Tien Nguyen\IEEEauthorrefmark{5}, 
        Dinh C. Nguyen\IEEEauthorrefmark{3},
        Aryan Kaushik\IEEEauthorrefmark{6},
        Periklis Chatzimisios\IEEEauthorrefmark{7}
    }
    \IEEEauthorblockA{
        \IEEEauthorrefmark{1}Faculty of Information Technology, Hung Yen University of Technology and Education, Hungyen, Vietnam \\
        \IEEEauthorrefmark{2}College of Sciences and Engineering, University of Tasmania, Sandy Bay, Hobart TAS 7001, Australia \\
        \IEEEauthorrefmark{3}Department of Electrical and Computer Engineering, University of Alabama in Huntsville, USA \\
        \IEEEauthorrefmark{5}Department of Electrical and Electronics Engineering, Lac Hong University, Vietnam \\
        \IEEEauthorrefmark{6}Department of Computing \& Mathematics, Manchester Metropolitan University, UK \\
        \IEEEauthorrefmark{7}Department of Information and Electronic Systems Engineering, International Hellenic University, Thessaloniki, Greece \\
    }
}

\begin{document}
\maketitle
\begin{abstract}
6G wireless networks are expected to provide seamless and data-based connections that cover space-air-ground and underwater networks. As a core partition of future 6G networks, Space-Air-Ground Integrated Networks (SAGIN) have been envisioned to provide countless real-time intelligent applications. To realize this, promoting AI techniques into SAGIN is an inevitable trend. Due to the distributed and heterogeneous architecture of SAGIN, federated learning (FL) and then quantum FL are emerging AI model training techniques for enabling future privacy-enhanced and computation-efficient SAGINs.  In this work, we explore the vision of using FL/QFL in SAGINs. We present a few representative applications enabled by the integration of FL and QFL in SAGINs. A case study of QFL over UAV networks is also given, showing the merit of quantum-enabled training approach over the conventional FL benchmark. Research challenges along with standardization for QFL adoption in future SAGINs are also highlighted.
\end{abstract}

\IEEEpeerreviewmaketitle

\section{Introduction}
In the past decades, terrestrial wireless communication systems have developed explosively, making distinguished contributions and setting the foundation for the development of human civilization. The development of mobile communication systems has gone through 5 generations, from 1G in the 1980s to 5G in the early 2020s. However, as an inevitable development trend, 6G is expected to be launched in the 2030s to provide extreme data rate network services and realize real-time IIoT applications \cite{9509294}. The architecture of 6G will completely cover Space-Air-Ground and underwater integrated networks \cite{10443704}. 
As a core partition of future 6G networks, Space-Air-Ground Integrated Networks (SAGIN) have been envisioned to provide countless real-time intelligent applications. To realize this, promoting AI techniques into SAGIN is an inevitable trend. Due to the distributed and heterogeneous architecture of SAGIN, FL is emerging and becoming one of the most promising AI models for SAGIN. However, employing FL in SAGIN still faces some challenges.
First, due to the large size and complexity of SAGIN, a large number of devices are required to collaborate to train a common model using an extremely large dataset. Moreover, as the dataset size and the required
model complexity increase, traditional client devices may not be able to perform the local ML training efficiently. Therefore, developing efficient FL solutions capable of handling large datasets and complex ML models in SAGIN is significantly important. Second, client devices in SAGIN rely on radio transmissions, which can be subject to potential eavesdropping attacks due to long-distance transmission. The nature of radio transmission may allow malicious nodes to eavesdrop on wireless signals and obtain model gradients by brute force cracking, thus inferring the private information of clients. 
Moreover, handling big data and complex AI models and ensuring privacy and security in the long-distance transmission of AI models also represent significant challenges.
To address these issues, we propose integrating FL into the SAGIN architectural framework to optimize system resource utilization, reduce communication latency and energy consumption, and enhance privacy and security, thereby realizing real-time, smart, and green 6G SAGIN-based IIoT applications.

\begin{figure}[!t]
\centering
\includegraphics[width=8cm]{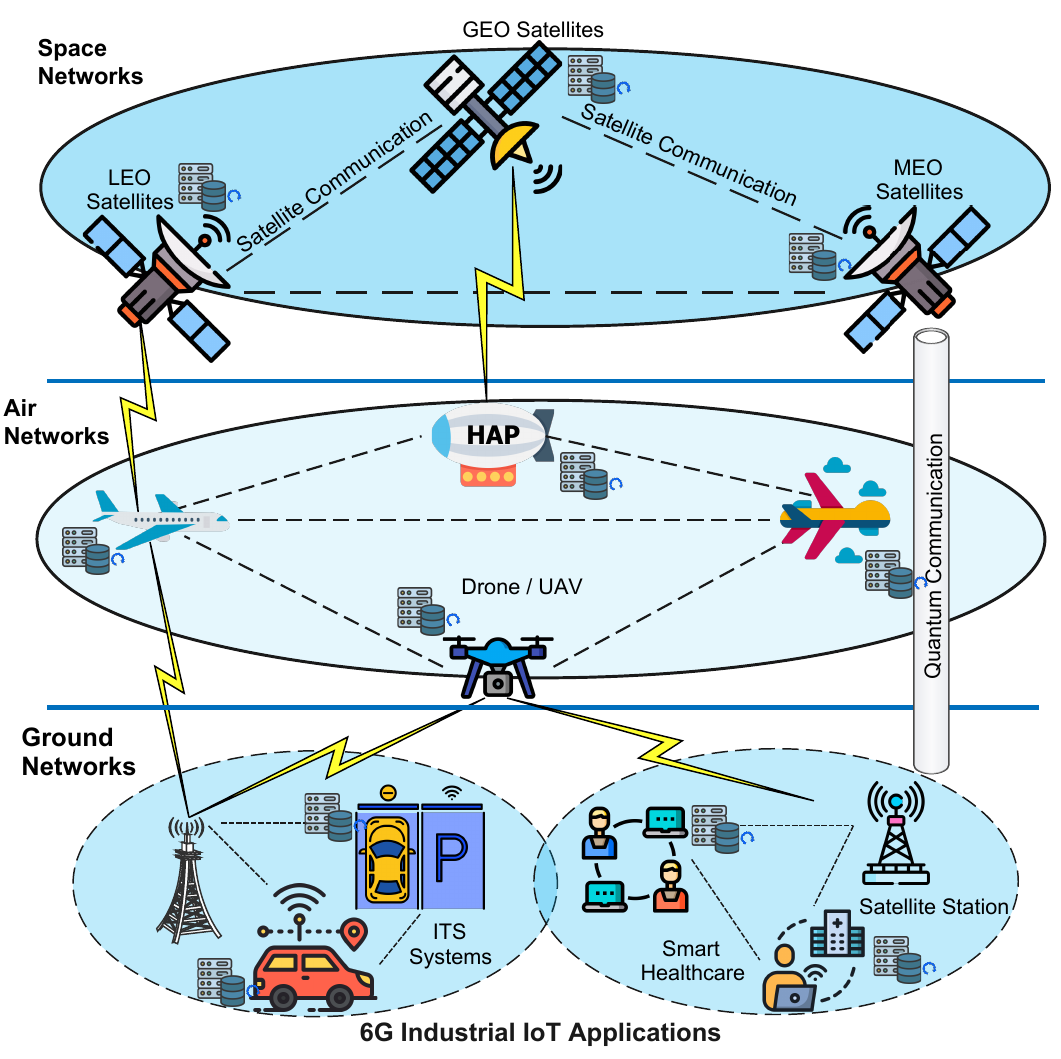}
\caption{An illustrated Architecture of FL-based SAGIN.}
\label{Fig1}
\end{figure}

\begin{figure}[!t]
\centering
\includegraphics[width=8cm]{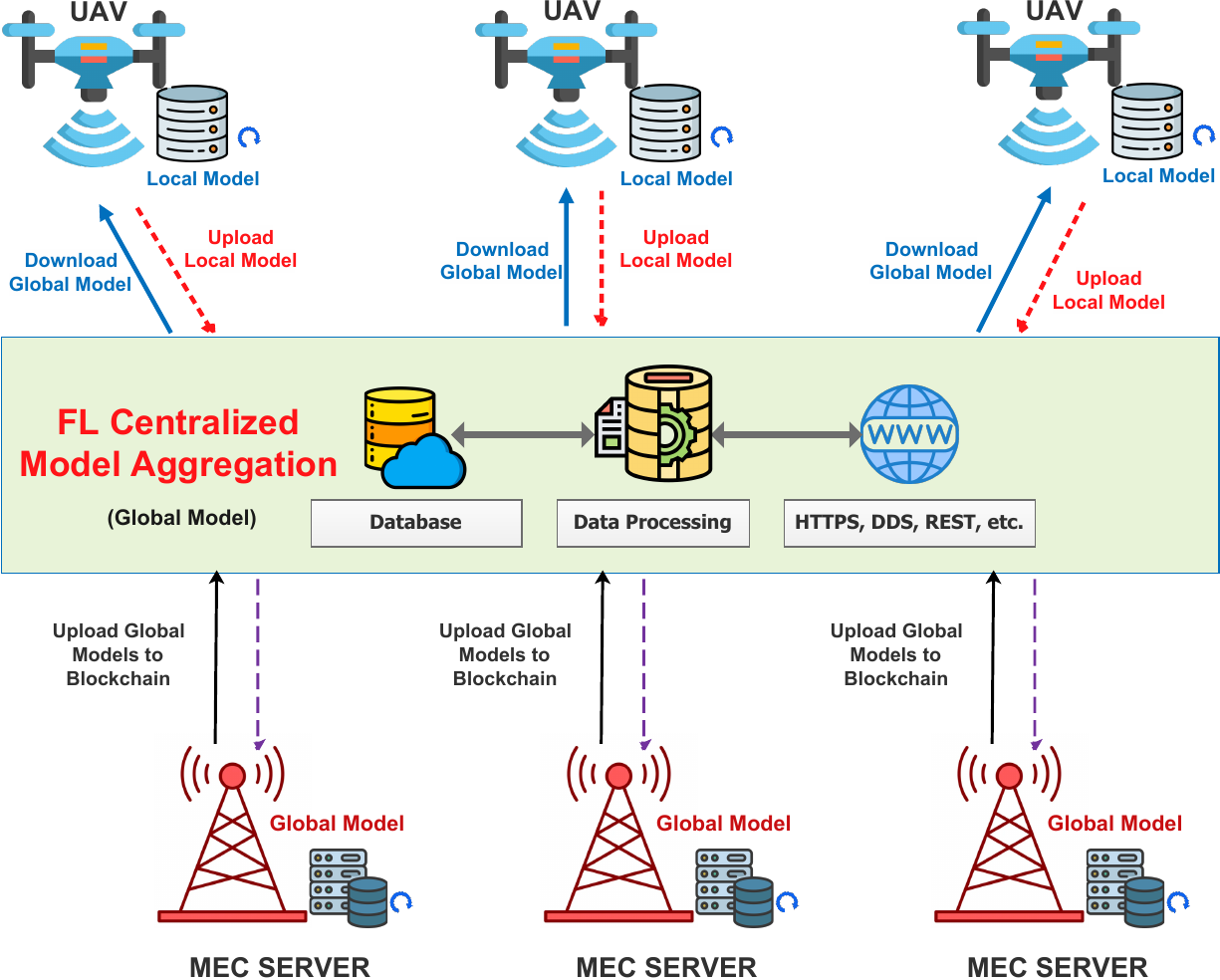}
\caption{An illustrated Architecture of FL-based UAV-Ground Networks.}
\label{Fig2}
\end{figure}

From a computational perspective, FL is a powerful distributed learning model that eliminates the limitations of traditional centralized learning models. The combination of FL and SAGIN architecture provides irreplaceable high-performance computing solutions in a series of domains such as military, UAV, ITS, real-time systems or complex mobile network architectures, where training efforts can be challenging for traditional computational techniques. Fig.~\ref{Fig1} presents a common FL-based SAGIN framework. After a finite number of training rounds, the system will converge, and the aggregator will send an optimal global model to IoT devices to ensure that the aggregator and all partners use the same training mode. Consequently, FL-based SAGIN provides a promising approach for these applications.
From the communications perspective, FL-based SAGIN systems are emerging as one of the promising 6G solutions to address the security and privacy challenges of traditional communications systems. Integrating FL and SAGIN architecture aims to take advantage of FL's unique characteristics, including distributed, security, and system resource optimization.
Recently, FL-based SAGIN frameworks have been focused on research. Fig.~\ref{Fig2} shows an FL-based UAV-Ground scheme. However, a comprehensive picture of enabling Space-Air-Ground Integrated Networks via FL has not been fully considered. 

Recently, due to the advances in quantum computing technologies, quantum FL (QFL) has emerged as a powerful solution for enabling next-generation SAGINs. QFL leverages the strengths of the distributed learning approach of FL and exponential speed enhancements characteristic of quantum computing. In SAGIN models, utilizing the satellite as the central FL server offers a promising strategy. However, employing FL in SAGIN faces challenges in managing vast datasets and training complex ML models. The work in \cite{10274653} proposes a quantum-enabled FL architecture for SAGIN that uses quantum relays and variational quantum algorithms (VQA). To effectively handle big datasets and intricate models within FL, the framework makes use of VQA-based machine learning for local training, all the while maintaining improved data security and privacy. It also includes a quantum relay technique to securely communicate machine learning models across large distances via quantum teleportation, supported by UAVs and HAPS. The practicality and effectiveness of this strategy are confirmed by numerical findings from a case study, which highlights the possibility of combining quantum technologies with FL in upcoming 6G networks. This paper presents a quantum-enabled FL architecture for SAGIN that makes use of quantum relays and VQA. To effectively handle big datasets and intricate models within FL, the framework makes use of VQA-based machine learning for local training, all the while maintaining improved data security and privacy. It also includes a quantum relay technique to securely communicate machine learning models across large distances via quantum teleportation, supported by UAVs and HAPS. The practicality and effectiveness of this strategy are confirmed by numerical findings from a case study, which highlights the possibility of combining quantum technologies with FL in upcoming 6G networks. The main contributions are summarized as follows.
\begin{itemize}
    \item We present recent advances in SAGINs and describe the new vision of FL and QFL in SAGINs, where the significance of privacy-preserving model training offered by FL and fast model training offered by QFL is highlighted. 
    \item We discuss a few representative applications enabled by integrating FL and QFL in SAGINs. A case study of QFL over UAV networks is also given.
    \item Future research challenges are also discussed, where standardization for QFL adoption in future SAGINs is highlighted.
\end{itemize}

\section{Recent Advances in SAGINs and FL}
\subsection{Space-Air-Ground Integrated Networks (SAGINs)}
According to the operation location, SAGINS's architecture includes three main network layers: Space, Aerial and Ground \cite{10398221}, as shown in Fig.~\ref{Fig1}.

- \textit{Satellites (Space Layer):} This layer includes satellites operating in different orbits. Based on operating orbit, the Space layer can be divided into 3 sublayers, including Low Earth Orbit (LEO), Medium Earth Orbit (MEO) and Geosynchronous Orbit (GEO). The LEO layer includes LEO satellites equipped with computing power and storage resources. Each satellite plays a role as a server with the capacity to process data, support edge computing and relay data to ground stations. It should be noted that LEO satellites move very quickly around the Earth so communication between ground devices and LEO satellites is not always available. Besides, LEO satellites are resource-constrained, so offloading strategies need to consider the computational task completion time of LEO satellites. Meanwhile, the MEO and GEO layers play the main role in relaying raw data to LEO satellites and ground stations.


- \textit{Air Layer:} This layer consists of high-altitude platforms (HAP) such as UAVs that can offer broadband wireless connections and real-time edge computing services. Moreover, HAPs support offloading tasks and caching services to reduce the end-to-end delay and energy consumption for ground devices. However, the HAP's coverage and energy supply are relatively limited.

- \textit{Ground Layer}
This layer includes all infrastructure and IoT network platforms deployed on the ground, such as base stations, sensors, vehicles, and ground computing centers. This layer collects data from devices and sensors to transceiver and execute calculated results from the above layers.

\subsection{Federated Learning}
FL is a distributed AI training model. Instead of centralized AI training on high-performance servers, FL allows devices to participate in training on its local datasets and only exchange training models with the server that has the role of an aggregator. Thanks to this, FL enhances privacy and security and realizes real-time IoT applications \cite{9475501}. FL consists of three main steps as follows.
\begin{itemize}
    \item \textbf{Initialization}:
The server acts as an aggregator that initializes the computational tasks, the learning parameter set (known as the $w^0_{G}$ model), the learning rate and the number of communication rounds. Then, it sends parameters to the training partners.
    \item \textbf{Local Traning}: The device $i^{th}$ uses the received $w^0_{G}$ model and its local dataset to train the more optimal updated model (known as the $w_{i}$ model) such as stochastic gradient descent. Then, the parameters of the updated model are resent to the aggregator.
    \item \textbf{Training Aggregator}: Then the aggregator receives all the updated models of the clients. It uses methods such as simple averaging or joint averaging to form a more optimal global model (known as the $w^i_{G}$ model). Then, the aggregator will send it back to the IoT devices for the next training round. 
\end{itemize}


\subsection{Quantum FL}
{Quantum Federated Learning (QFL) with quantum networks is a cutting-edge approach that combines quantum computing and federated learning to enhance distributed machine learning. In QFL, quantum networks enable multiple parties to collaboratively train a model without sharing their local data, maintaining privacy. In QFL, instead of traditional data processing, quantum data is used to train machine learning models across distributed quantum devices. Each participating device processes its local quantum data and shares only model updates, preserving the privacy of sensitive information. Quantum algorithms process and exchange information, potentially offering significant speedups and improved learning performance over classical methods. Using quantum entanglement in the networks ensures secure communication between the parties involved. This approach holds promise for solving complex problems in domains such as autonomous vehicles, smart healthcare, and 6G ecosystems \cite{10684196}.
In order to improve security and privacy within the Internet of Things ecosystem, this survey article looks at how federated learning, quantum computing, and 6G wireless networks can be integrated. The research in \cite{javeed2024quantum} investigates how federated learning enables IoT devices to cooperatively train a model without exchanging data, protecting privacy, and quantum computing can enhance encryption for more secure IoT data. IoT device connectivity and real-time secure data processing are improved by combining 6G's low latency and high speed.
Within a QFL framework, this work in \cite{qi2024federated} presents an effective federated quantum natural gradient descent (FQNGD) algorithm that uses VQC-based QNN. By minimizing the number of training iterations needed for convergence, the FQNGD algorithm lowers the overall communication cost among local quantum devices, hence minimizing the communication overhead. 


\subsection{Vision of Integrating FL into SAGIN}
The advent of 6G networks allows the provision of network services with extremely high throughput, leading to the explosive growth of real-time smart applications such as remote surgery, autonomous vehicles, augmented reality, and Metaverse \cite{10183802}. These systems generate huge amounts of data that require real-time computational processing. However, terrestrial wireless networks only cover about 20\% of the Earth's surface and under 6\% of the Earth's area. To address this issue, the study in \cite{9749193} proposed a satellite-based terrestrial communication system, namely SatCom. The authors claimed that SatCom is a viable solution for providing seamless network services based on the satellite's capacity, such as high data rate and global coverage. 
Furthermore, deploying wireless communication systems based on the support of UAVs is also a feasible solution. Thanks to the ability to be flexible and high-speed in moving and adapt to any terrain, UAVs can play a role as base stations to improve performance and extend coverage and energy efficiency. On the other hand, UAVs can serve as terminals in military, agriculture, and disaster response. However, due to resource-constrained UAVs, centralized training deployment of traditional DL techniques is unfeasible. Therefore, the work performed a comprehensive survey on integrating FDL into UAV-based wireless networks. This research shows that this integration can enhance privacy and security, and improve the performance and computational costs of UAVs-enabled wireless networks.
Moreover, QFL offers fast model training for SAGINs due to quantum computing. For example, skimmable quantum neural networks (sQNN)-based QFL can enhance model training rates over satellite-ground communications \cite{park2024dynamic}. These networks provide flexible configurations, such as angle and pole, which improve the adaptability of quantum computing. The framework dramatically improves communication capabilities by integrating successive decoding and superposition coding to optimize communication. 



\section{Applications of FL and QFL in SAGINs}
In this section, we highlight the ability to integrate FL in typical domains such as military operations, space networks, ground Internet of Things, and emergency response applications, as follows.

\subsection{FL-based SAGINs in Military Operations}
According to the military approach, FL is applied in most different military applications such as space, aerial, ground, and underwater.
The study in \cite{ref17} proposed a novel FL-based military-area classification framework. Indeed, aims to enhance privacy and security and the accuracy of aerial image identification, they apply FL to classify images collected from UAVs on both IID (Identically distributed) and non-IID (non-identically distributed) datasets. By leveraging FL, the system allows collaborative learning across multiple military organizations without sharing raw data, ensuring data privacy and security. The experimental results show that the proposed solution achieved a 100\% accuracy rate while enhancing privacy and security compared to existing solutions.
Also, in this direction, the work in \cite{ref18} indicated the disadvantages of the defense systems and proposed a novel computing framework based on integrating FL and blockchain for defense IoBT networks. The simulation results show that the proposed scheme improves significantly in terms of accuracy and loss rates while enhancing privacy and security compared to existing solutions.
Although FL has many advantages,  the synchronous convergence rate is one of the main challenges due to distributed training on heterogeneous devices. To address this issue, the research in \cite{10226108} proposed a scalable asynchronous federated learning scheme for civilian and military mission-critical IoT applications to enhance the convergence rate and privacy and security for real-time monitoring applications. The results showed that the proposed solution improved performance, computing costs and security compared to existing approaches.

\subsection{FL-based SAGINs in Space Applications}
Space networks are one of the important components of 6G architecture. Satellites can provide seamless network services for extremely large geographical areas and support terrestrial networks. However, orbit-based mobility and frequent disconnection leads to satellite-based communication is a real challenge. To address this issue, the study in \cite{10409275} proposed an FL-based satellite clustering method for intra-orbit inter-satellite links (ISL). Indeed, by predicting satellite visits, the proposed algorithm selects ISLs to mitigate the impact of disconnection. The results show that the proposed method improved significantly in terms of convergence speed and communication load to 7 and 10 times, respectively, compared to the existing strategy.
The security issues of integrating IoT devices into terrestrial and interplanetary networks, notably in space colonies. We propose an SDN-based FL system to improve satellite-IoT data protection. This system protects IoT data transmission with an SDN backbone, traffic regulator, and virtual firewall that categorises and blocks harmful traffic. Our implementation uses OpenMined-based federated learning to detect attacks with 79.47\% accuracy, plus differential privacy approaches to improve accuracy. Our approach also uses SVM, Decision Tree, and Random Forest, with Random Forest being the most accurate. This integration shows how FL can protect network security and data privacy in space colonization \cite{uddin2023sdn}.


\subsection{FL-based SAGINs in Emergency Response Applications} 
In emergency response scenarios such as earthquakes and tsunamis, terrestrial networks are unable to operate properly due to lost connectivity. SAGIN is one of the most viable solutions for establishing communication systems for search, rescue or environmental monitoring. To address this issue, the work in \cite{ACM21} proposed a communication network relying on 6G UAVs, FL, and blockchain for emergency response. To reduce end-to-end delay and energy consumption and enhance security, they estimated the forking event probability and expected energy consumption. The results show that the proposed method minimizes total energy consumption and optimizes smart disaster response systems.
In order to solve the issues of real-time data processing and privacy preservation, this research \cite{nguyen2021federated} presents a conceptual framework for smart crowd management in urban areas, utilising edge-assisted federated learning (FL). The suggested framework consists of four layers: application layers, edge computing, cloud computing, and data collecting. The approach improves privacy and security by using federated learning for training and locally processing data on edge devices to reduce risks such as label poisoning attacks in FL-based anomaly detection. Along with the framework's possible answers and practical ramifications, issues including device heterogeneity, user acceptance, scalability, and statutory compliance are explored. This approach has been shown to be effective in controlling big crowds and maintaining the safety and smooth flow of people and transportation in smart city settings through a case study and simulation results involving three to five local clients.

\subsection{QFL in SAGINs}
Quantum-based machine learning is emerging as a promising solution for advanced surveillance systems, offering enhanced capabilities in decision-making and task optimization in complex environments \cite{paul2024quantum}. QFL in SAGINs combines quantum computing, federated learning, and Space-Air-Ground Integrated Networks \cite{10274653}. SAGINs provide a vast and dynamic environment where data from satellites, aerial vehicles, and ground stations can be integrated for machine learning tasks. In QFL, the distributed architecture of SAGINs allows quantum nodes at different layers to collaboratively train models without sharing raw data, enhancing privacy and security. Quantum algorithms in this context can offer improved processing speeds and resilience to network delays or interference. Using quantum entanglement between nodes ensures secure and efficient communication across network layers. QFL in SAGINs has the potential to address complex challenges in areas like global surveillance, disaster management, and space exploration. It could also optimize resource allocation and improve decision-making in real-time across vast geographical regions. The convergence of quantum computing with SAGINs can revolutionize data processing and intelligence in highly distributed, heterogeneous 6G networks.
Here, This research \cite{han2024orchestrating} introduces a new QFL approach for improving machine learning services in remote places with weak terrestrial connection infrastructure through the use of SAGINs. This method overcomes the limited computational power of ground devices and the lack of terrestrial base stations by using nodes in space and air layers as edge computing units and model aggregators. 

\section{Case Study}
In this section, we present a case study where we develop a QFL Framework over UAV networks.
\subsection{System Model and Working Concept}
 The framework leverages quantum computing to enhance federated learning, where UAVs collaborate to train a global model without sharing their local data. A base station (BS) acts as the central server, responsible for aggregating the locally trained models from each UAV. This decentralized approach ensures data privacy, as sensitive raw data remains stored on the individual UAVs, while the BS facilitates efficient global learning by performing weighted averaging of the model parameters. The integration of quantum circuits further enhances the learning capability by allowing the use of quantum operations, improving the model's performance and convergence speed. Specifically, a quantum circuit is incorporated as a core layer in each UAV's model, where quantum gates, such as RX, RY, and RZ rotations, are used to manipulate qubits within the quantum device, allowing the model to explore and learn from quantum states that cannot be accessed by classical methods. Controlled quantum operations like CRX enable entanglement between qubits, further enriching the model's ability to capture complex patterns in the data. The use of quantum entanglement allows for more sophisticated correlations between data points, giving the framework a distinct advantage over traditional federated learning methods.
\subsection{Parameter Setting and Simulation Results}
We conduct numerical experiments to compare the performance of QFL and classical FL, evaluating both in terms of accuracy and loss. \textit{For the QFL framework,} simulations are conducted using the MNIST dataset, which is preprocessed to be distributed across multiple clients, each representing a UAV. Specifically, the data processing begins with the encoder layer, which converts classical input data, such as MNIST images, into quantum states for further processing by the quantum layers. The model uses the torch quantum's GeneralEncoder to map the input data onto qubits. Each UAV model includes a quantum circuit as a core layer, utilizing quantum gates such as RX, RY, and RZ for qubit rotations and CRX for controlled operations between qubits. These quantum operations are applied across 4 qubits (wires). The Adam optimizer has been employed for both local and global model training, with a learning rate of 0.001. The global model is trained for 50 epochs, with each epoch incorporating federated learning from the UAV clients. Each UAV uses a batch size of 64 for local training. Negative Log Likelihood loss was used to measure classification performance. \textit{For the FL framework,} we perform simulations using the MNIST dataset with traditional machine learning techniques, where the models are trained in a federated setup relying solely on classical methods.

\begin{figure}[ht!]
    \centering
    \footnotesize
    \begin{subfigure}[t]{0.49\linewidth} 
        \centering
        \includegraphics[width=\linewidth]{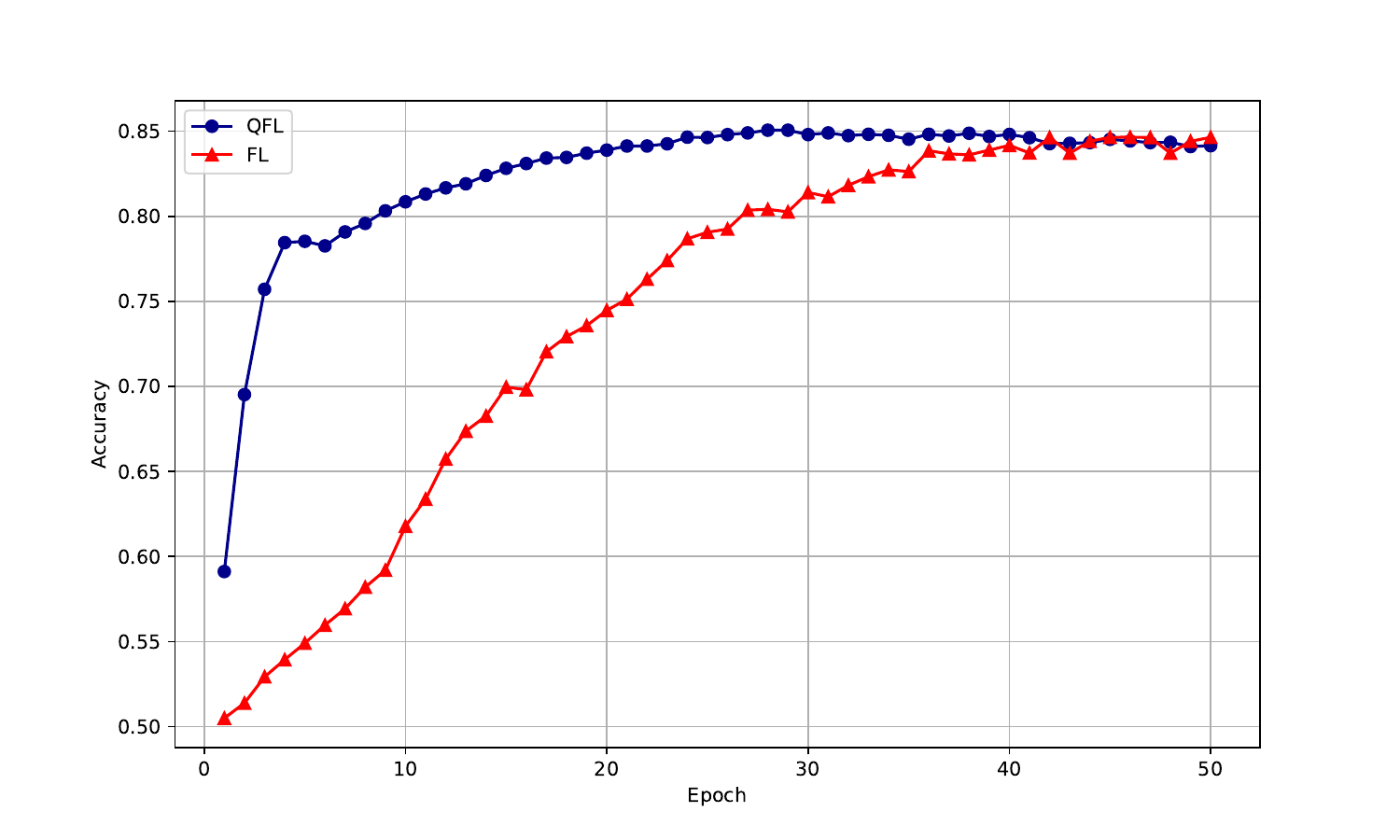}
        \caption{\footnotesize Comparison of accuracy between QFL and FL.}
        \label{fig3a}
    \end{subfigure}
    \hfill 
    \begin{subfigure}[t]{0.49\linewidth} 
        \centering
        \includegraphics[width=\linewidth]{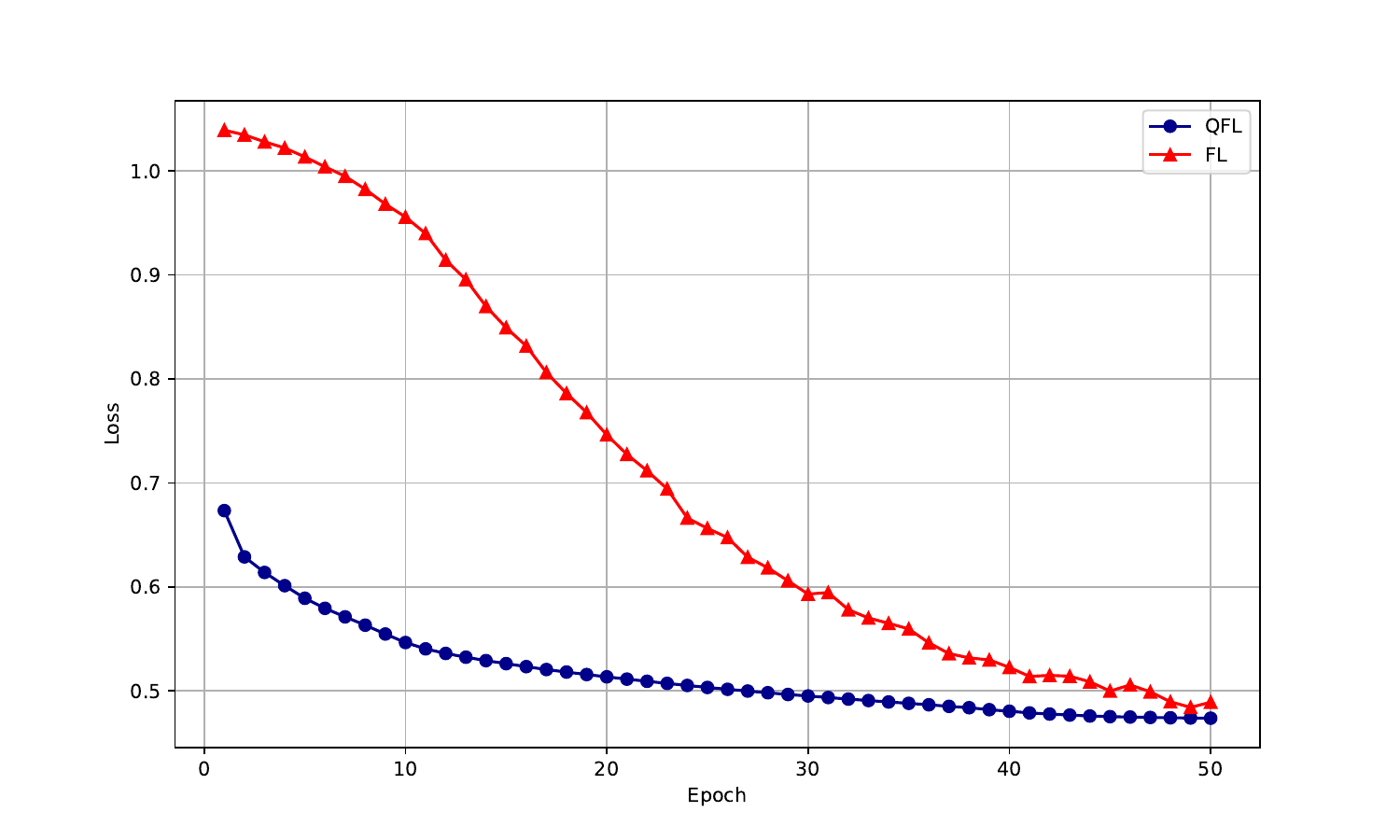}
        \caption{\footnotesize Comparison of loss between QFL and FL.}
        \label{fig3b}
    \end{subfigure}
    \caption{\footnotesize Comparison of QFL and FL in terms of accuracy and loss over the number of epochs.}
    \label{fig:accuracyloss}
\end{figure}

Fig.~\ref{fig:accuracyloss} illustrates a comparison of QFL and traditional FL in terms of accuracy and loss across a number of epochs. In Fig.~\ref{fig3a}, the accuracy of QFL and FL has been compared. Apparently, the QFL framework achieves higher accuracy at a faster rate, converging earlier than the classical FL technique. In Fig.~\ref{fig3b}, the loss comparison between QFL and FL indicates that QFL reduces loss more rapidly and stabilizes quicker than FL, further emphasizing the faster convergence of QFL. This faster convergence in QFL is primarily due to the integration of quantum circuits, which enables the exploration of richer state spaces and more efficient learning from complex data patterns. Quantum operations like entanglement and parallelism further enhance this capability, improving both accuracy and loss convergence.

\section{Challenges and Future Directions}
\subsection{Integrating 6G technologies in FL-based SAGINs}
The architectural characteristics of 6G are distributed and heterogeneous. Moreover, 6G technologies such as metaverse, digital twins, and quantum computing require low-extreme latency, energy efficiency, and rapid convergence. Meanwhile, the FL-based SAGIN architecture has extensive coverage, including space, ariel, and ground. Therefore, integrating 6G technologies in FL-based SAGINs is a significant challenge. In our view, intelligent elements based on endogenous AI are an important research direction. Besides the benefits of intelligent and customizable network elements deployed in space, aerial, and ground, from a design perspective, due to the extremely extensive coverage, the FL-based SAGINs architecture also requires research effective mechanisms, including routing, backup, and recovery paths and coordinated load balancing strategies \cite{9606832}.

\subsection{Security for FL-based SAGINs}
Privacy and security are the main challenges of the FL-based SAGINs architecture. From the application perspective, the expectations of the explosion of pervasive medical applications and autonomous vehicle systems in the 6G era can only be realized based on the guarantee of privacy and security. Although FL allows for distributed learning without central data storage on AI servers, SAGIN's extreme-massive coverage with space-aerial-ground collaboration services leads to it facing a series of security vulnerabilities, including personal IoT devices, gateways, edge and remote cloud servers. Furthermore, DDoS attacks can cripple IIoT systems, and data tampering attacks can lead to incorrect decisions by AI aggregators. In our vision, integrating blockchain technologies and lightweight encryption schemes are possible research directions for FL-based SAGINs \cite{9631953}.

\subsection{Resource Management for FL-based SAGINs}
With the complex architecture and vast coverage area of SAGINs, efficient resource management is a prerequisite for achieving seamless, reliable, and high-speed network services. Furthermore, next-generation networks will require complex handover services such as holographic communication with a huge number of mobile terminals. Therefore, efficient resource allocation and management schemes are vital to research directions for FL-based SAGINs to achieve a consistent and seamless perception of user services. Multiple intelligent resource management strategies are proposed. In our opinion, there are still some open challenges, including cross-layer resource allocation strategies and cloud-edge-end collaborative frameworks, task scheduling, and service-oriented network resource orchestration schemes \cite{10398221}. In reality, these are multi-constraint optimization problems with NP-hard complexity, so they pose significant challenges for FL-based SAGINs.
\subsection{Standardization for QFL}

Standardization for QFL is one of the vital issues in ensuring the seamless integration of quantum computing and federated learning across various platforms and industries. It involves defining protocols for communication between quantum devices, ensuring compatibility within diverse network environments, addressing data privacy and security, quantum model exchange formats, and the security of quantum communication channels, particularly with quantum cryptography techniques. Uniform guidelines for quantum algorithms, such as processing local quantum data and updating global models, will enhance efficiency and reliability. Another important aspect is defining error correction methods to handle quantum noise and decoherence, which are inherent challenges in quantum systems. Standardization allows scale-extended QFL applications across more extensive networks.
Moreover, international organizations must collaborate to establish legal and ethical frameworks for using QFL, ensuring data sovereignty and compliance with privacy laws. Standardized metrics for evaluating the performance and accuracy of QFL models can help benchmark their effectiveness. These standards will accelerate the adoption of QFL across sectors such as healthcare, finance, and telecommunications. Overall, comprehensive standardization is vital for creating robust, secure, and efficient QFL-based SAGINs.
The research in \cite{yun2022slimmable} shows a dynamic QFL architecture that combines QNNs with FL to address time-varying communication channels and constrained computational energy. SlimQFL uses the separable training of angle and pole parameters in QNNs to adapt to changing settings. SlimQFL achieves better classification accuracy than normal QFL, especially under poor channel circumstances, according to simulations. SlimQFL's convergence under different channel circumstances and non-IID data distributions could help determine quantum-enhanced federated learning systems' scalability and robustness.


\section{Conclusion}
In this work, we present a framework integrating FL and QFL into SAGINs. By addressing the challenges of big data, privacy, and security, the proposed FL-based SAGIN architecture optimizes system resource utilization, reduces communication latency, and enhances energy efficiency. Additionally, integrating Quantum Computing into FL to form Quantum Federated Learning (QFL) allows enhanced processing capacities and secure communication. Through this work, the effectiveness of the FL-based SAGIN framework is demonstrated, particularly in improving real-time IoT applications. This work highlights the potential of FL and QFL in transforming future 6G networks in several domains such as military operations and emergency response.

\bibliographystyle{IEEEtran}
\bibliography{References}

\end{document}